\documentclass{article}

\usepackage{microtype}
\usepackage{graphicx}
\usepackage{subcaption}
\usepackage{booktabs} 

\usepackage{hyperref}
\usepackage{xurl}
\usepackage{url}
\usepackage{multicol}
\usepackage{multirow}
\usepackage{enumitem}
\usepackage{fontawesome}
\usepackage{algorithm}

\usepackage[preprint]{icml2026}

\usepackage{amsmath}
\usepackage{amssymb}
\usepackage{mathtools}
\usepackage{amsthm}
\usepackage{float}
\usepackage{xspace}

\usepackage[capitalize,noabbrev]{cleveref}

\theoremstyle{pn}

\theoremstyle{definition}

\theoremstyle{remark}

\usepackage[textsize=tiny]{todonotes}

\begin{document}

\newcommand{\SysName}{$\mathtt{SpecForge}$\xspace}  
\newcommand{\BenchName}{$\mathtt{SpecBundle}$\xspace}

\twocolumn[
  \icmltitle{\SysName: A Flexible and Efficient Open-Source Training Framework for Speculative Decoding}
  


  \icmlsetsymbol{equal}{*}

  \begin{icmlauthorlist}
    \icmlauthor{Shenggui Li}{ntu,sf}
    \icmlauthor{Chao Wang}{mt,sf}
    \icmlauthor{Yikai Zhu}{sf}
    \icmlauthor{Yubo Wang}{sf}
    \icmlauthor{Fan Yin}{sf}
    \icmlauthor{Shuai Shi}{sf}
    \icmlauthor{Yefei Chen}{sf} \\
    \icmlauthor{Xiaomin Dong}{eigen}
    \icmlauthor{Qiaoling Chen}{ntu}
    \icmlauthor{Jin Pan}{sf}
    \icmlauthor{Ji Li}{sf}
    \icmlauthor{Laixin Xie}{sf}
    \icmlauthor{Yineng Zhang}{sf}
    \icmlauthor{Lei Yu}{mt}
    \icmlauthor{Yonggang Wen}{ntu}
    \icmlauthor{Ivor Tsang}{astar}
    \icmlauthor{Tianwei Zhang}{ntu}
  \end{icmlauthorlist}

  \icmlaffiliation{ntu}{Nanyang Technological University}
  \icmlaffiliation{sf}{SpecForge Team}
  \icmlaffiliation{mt}{Meituan}
  \icmlaffiliation{eigen}{EigenAI}
  \icmlaffiliation{astar}{A*STAR, Singapore}
    
  \icmlcorrespondingauthor{Tianwei Zhang}{tianwei.zhang@ntu.edu.sg}

  \icmlkeywords{Machine Learning, Machine Learning System, Infrastructure, Speculative Decoding}

  \vskip 0.1in
  \centering
  \text{\faGithub\  GitHub: \url{https://github.com/sgl-project/SpecForge}}
  
  \vskip 0.3in
]



\printAffiliationsAndNotice{}  

\begin{abstract}
Large language models (LLMs) incur high inference latency due to sequential autoregressive decoding. Speculative decoding alleviates this bottleneck by using a lightweight draft model to propose multiple tokens for batched verification. However, its adoption has been limited by the lack of high-quality draft models and scalable training infrastructure. We introduce \SysName, an open-source, production-oriented framework for training speculative decoding models with full support for EAGLE-3. \SysName incorporates target–draft decoupling, hybrid parallelism, optimized training kernels, and integration with production-grade inference engines, enabling up to 9.9$\times$ faster EAGLE-3 training for Qwen3-235B-A22B. In addition, we release \BenchName, a suite of production-grade EAGLE-3 draft models trained with \SysName for mainstream open-source LLMs. Through
a systematic study of speculative decoding training recipes, \BenchName addresses the scarcity of high-quality drafts in the community, and our draft models achieve up to 4.48$\times$ end-to-end inference speedup on SGLang, establishing \SysName as a practical foundation for real-world speculative decoding deployment.
\end{abstract}
\section{Introduction}

Large language models (LLMs) have rapidly become a cornerstone of modern AI systems. Both proprietary models—such as ChatGPT~\cite{Achiam2023GPT4TR}, Gemini~\cite{Reid2024Gemini1U, Comanici2025Gemini2P}, and Grok—and open-source counterparts, including LLaMA~\cite{Touvron2023LLaMAOA, Touvron2023Llama2O, Dubey2024TheL3}, DeepSeek~\cite{DeepSeekAI2024DeepSeekV3TR, DeepSeekAI2025DeepSeekR1IR}, and Qwen~\cite{Bai2023QwenTR, qwen2025qwen25technicalreport, yang2025qwen3technicalreport}, have driven substantial productivity gains across a wide range of industries. However, as model sizes continue to scale, inference latency has emerged as a fundamental bottleneck~\cite{Yu2022OrcaAD, recasens2025mindmemorygapunveiling}. LLMs' autoregressive generation requires a full forward pass through billions of parameters for each token, resulting in a memory-bound inference process that significantly increases deployment cost and hinders real-time or high-throughput applications.

Speculative decoding has emerged as a promising remedy, offering substantial speedups by pairing a small draft model with a large target model (e.g., the original model)~\cite{Leviathan2022FastIF, Chen2023AcceleratingLL}. The draft model quickly generates several candidate tokens, and the target model then verifies multiple tokens in parallel via a single forward pass. The draft model can be in the form of N-Gram models~\cite{Fu2024BreakTS}, models of smaller size from the same model family~\cite{Leviathan2022FastIF, Chen2023AcceleratingLL}, sub-layers of the same model~\cite{zhang-etal-2024-draft, Liu2024KangarooLS, Xia2024SWIFTOS}, and additional autoregressive adapters~\cite{Cai2024MedusaSL, Li2024EAGLESS, Li2024EAGLE2FI, Zhang2024LearningHR, Du2024GliDeWA, li2025eagle}. If the draft’s predictions are likely to be correct as judged by the target model, they are accepted; otherwise, the target model corrects them. The number of forward passes of the target model is thus significantly reduced. As a result, by effectively leveraging extra parallel computation when available, speculative decoding can reduce inference time substantially without changing the distribution of outputs.

Early demonstrations of speculative decoding~\cite{Leviathan2022FastIF} showed speedups of up to 3.4× on large Transformer models such as T5-XXL~\cite{10.5555/3455716.3455856}, while provably preserving output fidelity. Subsequent advances have further improved its efficiency and practicality. Notably, the EAGLE-3 algorithm~\cite{li2025eagle} introduces a draft model that operates at a hybrid feature level, substantially increasing token acceptance rates and achieving up to 4.79× speedup on LLaMA-3.3-70B without quality degradation. EAGLE-3 further incorporates dynamic tree-based generation and a Training-Time Test (TTT) procedure that better simulates multi-step decoding during draft training. Owing to its strong empirical performance, EAGLE-3 has become the de facto industrial standard for speculative decoding and is supported by major inference engines, including open-source systems such as SGLang~\cite{Zheng2023SGLangEE} and vLLM~\cite{kwon2023efficient}, as well as proprietary platforms like TensorRT-LLM~\cite{trtllm}.

However, despite the strong theoretical guarantees and empirical gains, speculative decoding remains underutilized in practice, especially for EAGLE3. We identify three main causes which hinder the wider application of speculative decoding in inference:

\textbf{Cause 1: Limited availability of draft models.} The effectiveness of speculative decoding critically depends on the adoption of a well-trained draft model that closely approximates the predictions of the target model. Such a draft model is often unavailable in practice. Early approaches~\cite{Leviathan2022FastIF} assume the existence of smaller models from the same family as the target model, which frequently does not hold. For example, Kimi K2~\cite{kimiteam2025kimik2openagentic}, a 1-trillion-parameter model, was released without a corresponding smaller variant, rendering this approach infeasible. Even with state-of-the-art methods such as EAGLE-3, many mainstream models, including Qwen3, lack publicly available matching draft models, significantly hindering the practical adoption of speculative decoding.

\textbf{Cause 2: Poor performance of open-source draft models.} Previous work~\cite{li2025eagle, Li2024EAGLESS} has released a limited number of draft-model weights on Hugging Face, enabling engineers and researchers to directly integrate them into inference frameworks such as SGLang, vLLM, and TensorRT-LLM for acceleration. However, these publicly available drafts are typically trained on relatively small, research-oriented datasets, which limits their robustness and renders them unsuitable for production-level deployment.

To quantify this limitation, we reproduced the EAGLE-3 training procedure for LLaMA-3.1-Instruct using ShareGPT (120K conversations) and UltraChat (200K conversations)~\cite{Ding2023EnhancingCL}. Under this setting, the resulting draft model achieved an acceptance length of 2.82 on the Math500 benchmark. In contrast, training the same draft model on the Perfect-Blend dataset, which contains 1.4M conversations, improved the acceptance length to 3.48, corresponding to an additional 1.17× inference speedup.

This gap highlights a broader mismatch in the open-source LLM ecosystem: while foundation models such as DeepSeek, Qwen, and GLM have rapidly advanced to state-of-the-art performance, high-quality draft models remain scarce and underdeveloped, leaving substantial headroom for improving the effectiveness of speculative decoding.

\textbf{Cause 3: Lack of robust training tools.} Constructing a high-quality draft model is inherently non-trivial, often requiring architectural customization and the implementation of sophisticated training mechanisms such as Training-Time Test (TTT)~\cite{li2025eagle}. Until recently, practitioners lacked robust tooling to support this process. Most existing speculative decoding implementations remain ad hoc, fragmented, or ill-suited for large-scale training~\cite{eagle-implementation}. Given that target models can range from several billion to over one trillion parameters, any practical training framework must be highly scalable, efficient, and reliable. The absence of such infrastructure has substantially hindered the community’s ability to train high-quality draft models, thereby limiting their availability and adoption across real-world and open-source ecosystems.

\par \SysName is our attempt to fill these gaps by advancing the practical development of speculative decoding in both research and industry. It is a unified, production-oriented framework for training draft models for speculative decoding, offering native support for advanced algorithms such as EAGLE-3, including the complex Training-Time Test (TTT) procedure with tree attention masks and recursive scheduling. With \SysName, practitioners can easily train state-of-the-art draft models through simple configuration rather than custom engineering.

To ensure that these capabilities operate at scale, \SysName adopts a \textbf{hybrid parallelism strategy via target-draft decoupling}, which explicitly decouples the target model and the draft model, enabling each to be parallelized according to its distinct computational characteristics. In speculative decoding training, the target model is typically large, frozen, and inference-dominated, while the draft model is lightweight and frequently updated. Treating both models as a single monolithic modules, as done in prior implementation, forces a uniform parallelization strategy that is suboptimal for both. \SysName instead applies inference-oriented parallelism to the target model \textbf{leveraging the integration with SGLang} and training-oriented strategies to the draft model. This separation improves scalability, reduces communication overhead, and allows the framework to efficiently support target models ranging from billions to over a trillion parameters.

\SysName further optimizes the Training-Time Test (TTT) procedure in EAGLE-3 by introducing memory- and compute-efficient attention implementations tailored to its autoregressive multi-step structure. It leverages the sparsity pattern in tree attention to reduce the computation time and memory peak, and optimizes the loss computation via customized in-place operations. Together, these optimizations significantly lower memory consumption and wall-clock time, enabling stable and scalable EAGLE-3 training at long context lengths.

To enrich the availability of high-quality draft models in the open ecosystem, we have trained a comprehensive suite of draft models, named \textbf{\BenchName}, covering mainstream open-source LLM families including Llama-3, Llama-4, Qwen-3, GPT-OSS, Kimi K2, and DeepSeek V3. \BenchName is built on extensive, diverse training corpora specifically for speculative decoding, offering substantially stronger draft quality than existing open-source checkpoints. In empirical evaluations, \BenchName models deliver up to 4.8x speedup over inference without speculative decoding and 1.3× speedup over publicly available draft model checkpoints across multiple task domains, making them practical drop-in draft models for real-world deployment.

During the training of \BenchName, we also systematically investigated the properties of speculative decoding and derived practical training recipes that elucidate key design choices, including draft model architectures, dataset quality, and the configuration of training-time test. Together, these findings provide actionable guidance for building high-quality draft models and inform best practices for deploying speculative decoding in production settings.

We summarize our main contributions and key novel features of \SysName as follows:
\begin{itemize}[leftmargin=*, itemsep=0pt, topsep=-7pt, parsep=0pt]
    \item We introduce \textbf{\SysName}, an efficient and scalable training framework for speculative decoding. \SysName implements hybrid parallelism through target–draft decoupling and training-time test attention optimization, enabling large-scale and efficient draft model training.
    \item We release \textbf{\BenchName}, a collection of high-quality draft models covering mainstream open-source LLM families. It delivers stronger accuracy and up to 1.3$\times$ speedup over existing open checkpoints, addressing the limited availability of draft models in the open-source ecosystem.
    \item We systematically investigate training recipes and design choices for improving speculative decoding performance, providing practical insights for real-world deployment and directions for future research.
\end{itemize}
\section{Preliminaries}

\subsection{Speculative Decoding}

Speculative decoding ~\cite{Chen2023AcceleratingLL,Leviathan2022FastIF,xia2023speculative,10.1145/3620666.3651335,Cai2024MedusaSL,Li2024EAGLESS,Li2024EAGLE2FI,li2025eagle,hu2025speculative} has emerged as the premier algorithmic intervention to address this memory-bound inefficiency without requiring the retraining of the foundational model or compromising generation quality. First formalized effectively by ~\citet{Leviathan2022FastIF}, it fundamentally restructures the inference workload. It replaces the serial, memory-intensive generation of single tokens with a parallel, compute-intensive verification of candidate sequences. By employing a computationally inexpensive ``draft model'' to propose short sequences of tokens (speculation), speculative decoding allows the massive ``target model'' to verify these proposals in a single forward pass. This effectively converts the sequential generation problem into a batch processing problem, thereby increasing arithmetic intensity and better utilizing the massive parallel compute capabilities of modern hardware.

\subsection{Theoretical Speedup}

The efficiency of speculative decoding is governed by the trade-off between the time saved by accepting draft tokens and the overhead of generating them. The number of tokens generated per single run of the target model is a random variable dependent on the quality of the draft model. Assuming the acceptance of each token is an independent event with probability $\alpha$ (the acceptance rate), the expected number of tokens generated per cycle is derived from a truncated geometric distribution:
$$E[\text{tokens}] = \frac{1 - \alpha^{\gamma+1}}{1 - \alpha}$$
where $\gamma$ is the number of speculative draft tokens. As $\alpha$ approaches 1 (perfect alignment between draft and target), the expected length approaches $\gamma + 1$. The theoretical walltime speedup $S$ is defined as the ratio of standard autoregressive time to speculative decoding time. Let $c$ be the cost ratio between the draft model and the target model ($c = C_q / C_p$). The speedup is given by:
$$S = \frac{E[\text{tokens}]}{1 + \gamma c} = \frac{1 - \alpha^{\gamma+1}}{(1 - \alpha)(1 + \gamma c)}$$
This equation reveals the critical efficiency bounds: 
\begin{itemize}[leftmargin=*, itemsep=0pt, topsep=-7pt, parsep=0pt]
    \item Draft Quality ($\alpha$): Maximizing $\alpha$ is paramount. The acceptance rate is fundamentally limited by the Kullback-Leibler (KL) divergence between the draft and target distributions.
    \item Draft Cost ($c$): The draft model must be significantly cheaper than the target ($c \ll 1$). If the draft overhead $\gamma c$ becomes too large, it negates the parallelization gains.
    \item Speculation Length ($\gamma$): There is an optimal $\gamma$ for any given $\alpha$ and $c$. While increasing $\gamma$ raises the potential tokens per step, it linearly increases the draft overhead. Modern frameworks often tune $\gamma$ dynamically.
\end{itemize}

\subsection{Architecture Paradigm}

The draft model in speculative decoding has undergone a series of paradigm shift. 

\textbf{Stage 1: Independent Draft Model}. The earliest realizations of speculative decoding adopted a straightforward architectural strategy in which a smaller, independently trained language model serves as the draft model~\cite{Leviathan2022FastIF, Chen2023AcceleratingLL, NEURIPS2023_6034a661, NEURIPS2023_7b97adea}. While conceptually simple, this paradigm imposes significant system-level constraints. To preserve tokenizer compatibility, a large target model (e.g., Chinchilla-70B or LLaMA-2-70B) must be paired with a substantially smaller model, often from the same family (e.g., 7B variants). This tight architectural coupling introduces an inherent alignment gap: smaller models tend to produce probability distributions that diverge from those of their larger counterparts, particularly on complex reasoning or long-context tasks, resulting in elevated token rejection rates. From the system perspective, maintaining two independent models also increases memory pressure, as both model parameters and KV caches must reside in GPU memory. In distributed settings, this design further incurs synchronization and communication overhead, which can erode the theoretical speedups of speculative decoding.

\textbf{Stage 2: Multi-Token Prediction Heads}. Medusa~\cite{Cai2024MedusaSL} introduced multiple prediction heads to eliminate the overhead of independent draft models and dual KV-caches. It adds lightweight heads that run in parallel with the standard LM head, enabling zero-latency drafting: generating candidate tokens costs nearly the same walltime as generating one because the additional MLPs are small and executed concurrently with the backbone. These heads reuse the target model’s features, avoiding a separate KV-cache and ensuring strong alignment. Verification is performed in a single forward pass using Tree Attention~\cite{Cai2024MedusaSL}, where a masked attention structure constrains each candidate token to attend only to its ancestors, allowing the model to evaluate multiple hypotheses in parallel and preserve useful branches even when others are rejected.

\textbf{Stage 3: Feature-Level Extrapolation}. Although Medusa eliminates the overhead of maintaining an independent draft model, its non-autoregressive MLP heads struggle to capture long-range dependencies. \citet{Li2024EAGLESS} address this limitation with EAGLE, which shifts autoregression from token space to feature space under the feature-uncertainty hypothesis: hidden-state trajectories in high-dimensional feature space are smoother and more predictable than the discrete jumps between tokens~\cite{Li2024EAGLESS, Du2024GliDeWA}. EAGLE replaces the standalone draft model with a lightweight single-layer Transformer that autoregressively predicts future feature representations, which are then projected through a linear layer to obtain the token logits. This fully autoregressive yet efficient design enables accurate multi-step drafting and achieves substantial empirical speedups. Its successor, EAGLE-2~\cite{Li2024EAGLE2FI}, further improves performance by dynamically shaping the draft tree according to token-level confidence, allocating verification compute to the most promising candidates.

\begin{algorithm}[H]
\caption{TTT Attention}
\label{algo:naive-ttt}
\begin{algorithmic}
\STATE {\bfseries Input:} query $q_t$, prefix keys $K^{\mathrm{train}}$, prefix values $V^{\mathrm{train}}$, cached keys $\{k_i\}_{i>T}$, cached values $\{v_i\}_{i>T}$
\STATE {\bfseries Output:} attention output $o_t$
\STATE

\STATE $S \gets \frac{q_t \left(K^{\mathrm{train}}\right)^{\top}}{\sqrt{d_k}}$

\FOR{$i \gets T+1$ to $t-1$}
    \STATE $s_i \gets \frac{q_t \cdot k_i}{\sqrt{d_k}}$
    \STATE $S \gets \mathrm{concat}(S,\ s_i)$
\ENDFOR

\STATE $\alpha \gets \mathrm{softmax}(S)$
\STATE $o_t \gets \alpha \cdot V^{\mathrm{train}}$

\FOR{$i \gets T+1$ to $t-1$}
    \STATE $o_t \gets o_t + \alpha_i\, v_i$
\ENDFOR

\end{algorithmic}
\end{algorithm}

\subsection{Training-Time Test}

In the latest upgrade, EAGLE-3~\cite{li2025eagle} additionally incorporates  \emph{Training-Time Testing (TTT)} to autoregressively generate the next few tokens, reducing error accumulation and improving acceptance rates in multi-token prediction. The core idea of TTT is to simulate multiple steps of autoregressive token generation during training. As shown in Algorithm \ref{algo:naive-ttt}, at each TTT step, the model attends to a growing context consisting of the original training sequence as prefix and the tokens generated in previous steps.

Let \(T\) denote the length of the training prefix. For any position \(t > T\), positions \(1{:}T\) correspond to the training sequence, while positions \(T{+}1{:}t{-}1\) correspond to representations predicted during earlier TTT steps. At position \(t\), the model computes attention using the query vector \(q_t\) over keys and values concatenated.

We define the key and value matrices for the training prefix and previously
predicted tokens as
\[
K^{\mathrm{train}} = [k_1,\ldots,k_T], \quad
K^{\mathrm{pred}} = [k_{T+1},\ldots,k_{t-1}],
\]
\[
V^{\mathrm{train}} = [v_1,\ldots,v_T], \quad
V^{\mathrm{pred}} = [v_{T+1},\ldots,v_{t-1}].
\]

The attention output at step \(t\) is computed as
\[
o_t
=
\mathrm{softmax}\!\left(
\frac{
q_t
\begin{bmatrix}
K^{\mathrm{train}} \\
K^{\mathrm{pred}}
\end{bmatrix}^{\!\top}
}{\sqrt{d_k}}
\right)
\begin{bmatrix}
V^{\mathrm{train}} \\
V^{\mathrm{pred}}
\end{bmatrix},
\]
where \(d_k\) denotes the key dimensionality.

Equivalently, the attention logits can be decomposed into prefix and prediction
components as
\[
S_t
=
\left[
\frac{q_t {K^{\mathrm{train}}}^{\!\top}}{\sqrt{d_k}},
\;
\frac{q_t {K^{\mathrm{pred}}}^{\!\top}}{\sqrt{d_k}}
\right].
\]

Intuitively, the attention logits \(S_t\) decompose into two parts: (i) causal
attention between the query \(q_t\) and the training prefix keys
\(K^{\mathrm{train}}\), and (ii) dot products between \(q_t\) and the keys
\(k_i\) generated in previous TTT steps for \(i>T\):

\section{Challenges}

\par Despite the rapid growth of speculative decoding, especially EAGLE3, training the draft model has received less attention. Compared to large-scale model training using frameworks like Megatron~\cite{10.1145/3458817.3476209} and DeepSpeed~\cite{Rasley2020DeepSpeedSO}, one significant attribute of EAGLE3 training is that the number of trainable parameters is smaller by a magnitude as the EAGLE3 draft model is often one-layer Transformer. Nonetheless, constructing a draft model is non-trivial because of the following challenges.

\par \textbf{Rigid Parallelism Strategies.} Existing open-source implementations~\cite{eagle-implementation, modelopt} treat the target and draft model as a unified model, and apply fully sharded data parallelism (FSDP)~\cite{Zhao2023PyTorchFE, Rasley2020DeepSpeedSO} for training by wrapping both models. Despite its simplicity and user-friendliness, such unified parallelism strategy is sub-optimal for several reasons. First, even though the draft models are typically small, the target models can vary from several billions of parameters to trillions of parameters. ZeRO-style sharding~\cite{Rasley2020DeepSpeedSO} is not optimal for all scales and thus does not provide high-performance hidden-state generation. It is evident that current high-performance generation engines like SGLang, vLLM and TensorRT favour tensor parallelism and expert parallelism over all-gather-based ZeRO-style sharding. Consequently, treating the target and draft models as a single monolithic module limits both performance and scalability.

\par \textbf {Sub-optimal Prefill Performance.}
The training process of EAGLE3 can be naturally decomposed into two stages. In the first stage, the target model is executed over the entire input sequence to generate the corresponding hidden states. This is equivalent to the prefill phase in standard LLM inference, where the model processes the tokens in a fully autoregressive manner in parallel before decoding begins.

However, existing EAGLE3 training frameworks typically rely on naïve model implementations, either self-written or directly imported from Hugging Face. These implementations are primarily designed for general-purpose training and correctness, rather than for high-throughput inference workloads. As a result, they fail to exploit many inference-specific optimizations that have been extensively engineered into mature, production-grade inference engines.
In particular, these training pipelines lack optimizations such as efficient attention kernels, optimized memory management, and CUDA Graph of which are critical for accelerating the prefill stage. In contrast, modern inference engines like SGLang and vLLM are explicitly optimized for this execution pattern and can deliver substantially higher throughput and better hardware utilization during prefill.

This mismatch leads to a significant inefficiency in training: the prefill stage often becomes a dominant bottleneck in large-scale draft-model training, inflating both training time and resource consumption. Addressing this gap requires rethinking the training pipeline to better align with inference-optimized execution.

\begin{figure*}[t]
  \centering
  \includegraphics[width=0.95\linewidth]{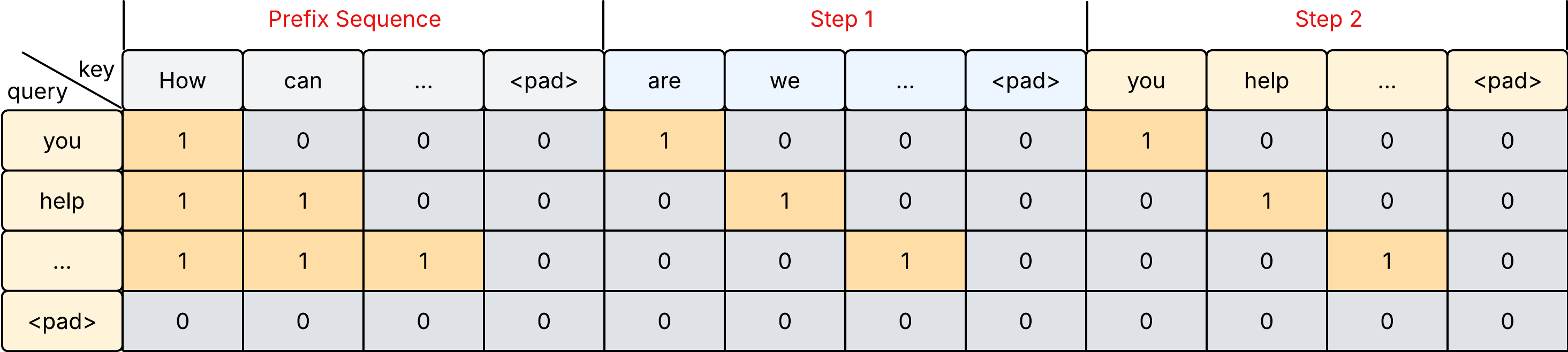}
  \caption{EAGLE3 attention mask used in Training-Time Testing.}
  \label{fig:eagle3-mask}
\end{figure*}

\section{\SysName}

\par We proposes several techniques to tackle the above challenges and optimize the overall training performance.

\begin{figure}[thb]
    \centering
    \begin{subfigure}{\linewidth}
        \centering
        \includegraphics[width=0.9\linewidth]{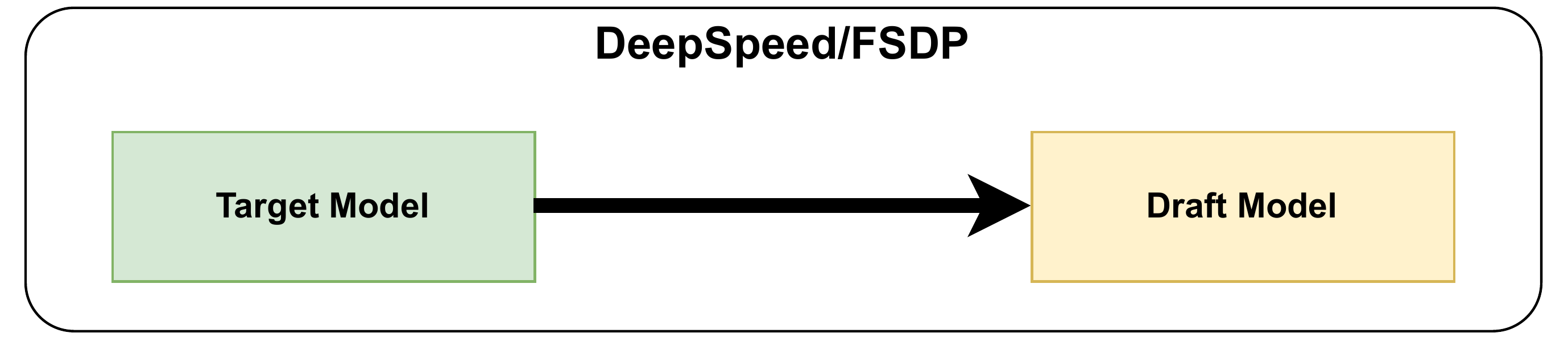}
        \caption{Existing implmenetation wraps both the target model and draft model into a single parallel strategy}
        \label{fig:existing-arch}
    \end{subfigure}

    \vspace{0.5em} 

    \begin{subfigure}{\linewidth}
        \centering
        \includegraphics[width=0.9\linewidth]{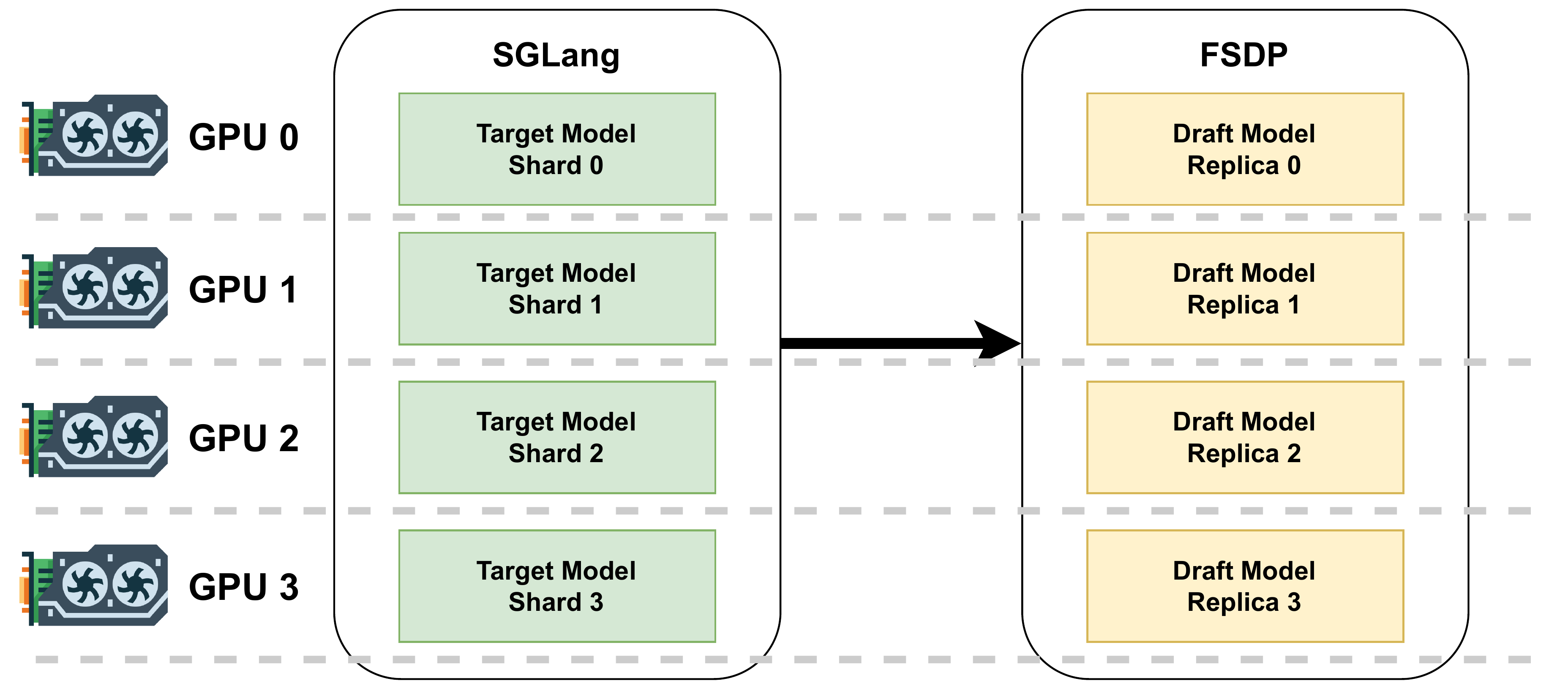}
        \caption{\SysName decouples the target model and draft model with hybrid parallelism}
        \label{fig:specforge-arch}
    \end{subfigure}

    \caption{Architecture comparisons}
    \label{fig:arch}
\end{figure}

\subsection{Target-Draft Decoupling}

The original EAGLE3 design tightly couples the draft and target models into a single parallelized module, as illustrated in Figure~\ref{fig:existing-arch}. While this design simplifies implementation, it is suboptimal from a performance standpoint. More critically, training and inference engines are optimized for fundamentally different objectives and system constraints; tightly coupling the two models prevents the simultaneous use of state-of-the-art training frameworks and high-performance inference engines. Decoupling the draft and target models therefore emerges as a key abstraction for achieving scalability, efficiency, and deployment flexibility.

For the draft model, the primary challenge lies in training efficiency which has been well supported by mature training frameworks such as DeepSpeed~\cite{Rasley2020DeepSpeedSO} and Megatron~\cite{10.1145/3458817.3476209}, offering extensive parallelization for distributed training. In contrast, the target model is typically large and inference-only, making it better suited to specialized inference engines such as SGLang~\cite{Zheng2023SGLangEE}. By decoupling the two models and applying distinct execution backends and parallelization strategies, \SysName enables each component to operate in its optimal regime. This also allows the direct deployment of trained draft models on optimized inference engines, resulting in a seamless and production-ready workflow.

\subsubsection{Hybrid Parallelism}

To accommodate the distinct characteristics of target and draft models, we leverage the SGLang inference engine and Fully Sharded Data Parallelism for them respectively, as shown in Figure~\ref{fig:specforge-arch}. 

For the draft model, this design choice is motivated by two key observations. First, the draft model is typically only 3–5\% the size of the target model, rendering heavyweight parallelization schemes such as tensor or pipeline parallelism unnecessary or even counterproductive. Second, training and inference stages impose fundamentally different compute and memory requirements, making specialized, stage-specific optimizations critical to overall system performance. Given the relatively small size of the draft model, we only shard the optimizer states and gradients to minimize the communication overhead, which is equivalent to ZeRO Stage 2 in DeepSpeed~\cite{Rasley2020DeepSpeedSO}.

For the target model, we directly employ the model runner provided by SGLang as the inference engine. This allows us to reuse SGLang’s existing parallelization strategies—including tensor parallelism, expert parallelism, and pipeline parallelism—as well as high-performance kernels such as FlashAttention~\cite{10.5555/3737916.3740109} and FlashInfer~\cite{Ye2025FlashInferEA} to accelerate the prefill phase of target-model inference. In addition, we can apply piecewise CUDA Graph in SGLang to fuse non-attention modules into a single kernel to reduce kernel launch time.

By decoupling the parallel strategies of the target model and draft model, they can be either co-located on the same GPU, or disaggregated on distinct GPUs. For our experiments, we conducted evaluation under the co-locate settings.

\begin{algorithm}[H]
\caption{BlockMask Construction for Training-Time Testing}
\label{alg:blockmask}
\begin{algorithmic}

\STATE {\bfseries Input:} batch index $b$, query index $q_i$, key/value index $kv_i$, prefix length $Q_{\text{LEN}}$, sequence length $T$
\STATE {\bfseries Output:} BlockMask $M$
\STATE

\STATE \textbf{// Causal mask}
\STATE $m_{\text{causal}} \leftarrow (q_i \ge kv_i)$
\STATE $m_{\text{pad}} \leftarrow (kv_i < T)$
\STATE $M_{\text{causal}} \leftarrow m_{\text{causal}} \land m_{\text{pad}}$

\STATE

\STATE \textbf{// Suffix mask}
\STATE $m_{\text{suffix}} \leftarrow (kv_i \ge Q_{\text{LEN}})$
\STATE $m_{\text{pad}} \leftarrow (kv_i \bmod Q_{\text{LEN}} < T)$
\STATE $m_{\text{diag}} \leftarrow ((kv_i - q_i) \bmod Q_{\text{LEN}} = 0)$
\STATE $M_{\text{suffix}} \leftarrow m_{\text{suffix}} \land m_{\text{pad}} \land m_{\text{diag}}$

\STATE

\STATE \textbf{// BlockMask}
\STATE $M \leftarrow M_{\text{causal}} \lor M_{\text{suffix}}$

\end{algorithmic}
\end{algorithm}

\subsection{Computation Optimization}

Beyond parallelization strategies, we further investigated the training characteristics of the draft model and observed that Training-Time Test (TTT) with a step length of 7 incurs substantial GPU memory consumption. To address this bottleneck, we design two complementary optimizations that significantly reduce memory usage during training.

\subsubsection{Sparse Tree Attention}

The naive implementation in Algorithm~\ref{algo:naive-ttt} materializes the attention logits S as intermediate activations. As TTT runs multiple autoregressive steps in the forward pass, these logits accumulate and quickly dominate memory usage. Our profiling shows that stored attention logits account for \textbf{80\%} of the total activation memory, making them the primary memory
bottleneck during training.

To reduce the memory footprint, we leverage \emph{FlexAttention} to compute attention. FlexAttention is a PyTorch project that leverages TorchInductor to compile a Python DSL into a Triton kernel. It brings two benefits:
(1) FlexAttention computes attention in a FlashAttention-style streaming
manner, avoiding the need to save intermediate activations; and
(2) FlexAttention implements a \emph{BlockMask} data structure, which efficiently precomputes blocks that can be skipped, partially computed, or fully computed, and optimizes the implementation accordingly. To use \emph{FlexAttention}, we construct a \emph{BlockMask} that encodes the allowed attention blocks, as illustrated in Figure~\ref{fig:eagle3-mask} and Algorithm \ref{alg:blockmask}. At each TTT step we store the newly generated keys and values in the KV cache and construct a custom attention mask represented as a \emph{BlockMask}, which is then provided to the FlexAttention operator during attention computation.

\subsubsection{Memory-Efficient Gradient Computation}

To further reduce memory usage, we implement the backward pass of the masked softmax loss with a custom Triton kernel (Algorithm \ref{alg:logsoftmax-backward}). The key idea is to reuse the input logits tensor to store gradients during the backward pass. After the forward loss computation, the logits are no longer needed. Instead of allocating a separate gradient buffer, the Triton kernel overwrites the logits tensor with the gradient with respect to logits. This avoids storing additional activation or gradient tensors and reduces memory overhead. The memory reduction ranges from \textbf{30--40\%}, depending on the context length and the draft model's vocabulary size.

\begin{algorithm}[H]
\caption{In-place Backward for Log-Softmax Loss}
\label{alg:logsoftmax-backward}
\begin{algorithmic}
\STATE {\bfseries Input:} logits $z$, target $p$, upstream gradient $g$
\STATE {\bfseries Output:} gradient w.r.t. logits (stored in $z$)

\STATE $s \leftarrow \sum (p \cdot g)$\;
\STATE

\STATE $\pi \leftarrow \mathrm{softmax}(z)$\;
\STATE $z \leftarrow - (p \cdot g - \pi \cdot s)$\;
\end{algorithmic}
\end{algorithm}
\section{Evaluation of \SysName}

\begin{table*}[t]
\centering
\resizebox{\linewidth}{!}{
\begin{tabular}{c|c|c|c|c|c|c|c|c}
\toprule[1.5pt]
\textbf{Target Model} & \textbf{Framework} & \textbf{Target Model} & \textbf{Draft Model} & \textbf{Max Batch Size} & \textbf{Seq Length} & \textbf{Step Time (s)} & \textbf{Throughput (tokens/s)} & \textbf{speedup} \\ \midrule[1pt]

\multirow{2}*{Llama3.1-8B}  & EAGLE & ZeRO 2 & ZeRO 2  & 16 & \multirow{2}*{4096} & 1.04 & 63015.4  & 1     \\
                             & \SysName & TP=1   & ZeRO 2  & 64 &  & 2.07 & 126639.6 & 2.01  \\ \hline

\multirow{3}*{Llama3.3-70B} & EAGLE & ZeRO 2 & ZeRO 2 & 16 & \multirow{3}*{4096} & OOM  & -        & -   \\
                             & EAGLE & ZeRO 3 & ZeRO 3 & 8  &  & 2.21 & 14827.1  & 1   \\                                
                             & \SysName & TP=4   & ZeRO 2 & 16 &  & 3.18 & 20608.8  & 1.39    \\ \hline

\multirow{3}*{Qwen3-30B-A3B} & EAGLE & ZeRO 2 & ZeRO 2 & 8  & \multirow{3}*{4096} & 1.12 & 29257.1  & 1    \\
                             & EAGLE & ZeRO 3 & ZeRO 3 & 8  &  & 5.07 & 6463.1   & 0.2    \\
                             & \SysName & TP=4   & ZeRO 2 & 16 &  & 0.52 & 126030.8 & 4.31     \\ \hline

\multirow{3}*{Qwen3-235B-A22B}  & EAGLE & ZeRO 2 & ZeRO 2 & 8  & \multirow{3}*{4096} & OOM  & -        &  -    \\
                             & EAGLE & ZeRO 3 & ZeRO 3 & 8  &  & 11.2 & 2025.7   &  1    \\
                             & \SysName & TP=8   & ZeRO 2 & 8  &  & 1.62 & 20227.2  &  9.99    \\ \bottomrule[1.5pt]
\end{tabular}}
\caption{End-to-end performance on various models.}
\label{tab:e2e-perf}
\end{table*}


\subsection{Experimental Setup}

We evaluated the system performance of \SysName and compare it against existing implementations. We considered two publicly available codebases: (1) the official implementation released by SafeAILab alongside the EAGLE3 paper, and (2) a third-party implementation developed by NVIDIA’s Model Optimizer team. As both implementations adopt a similar monolithic architecture wrapping both the target and draft models within DeepSpeed, we select the official SafeAILab implementation as our baseline. All experiments were conducted on a cluster of eight NVIDIA H200 GPUs with a sequence length of 4096. The batch size was adjusted for each method to maximize throughput under GPU memory constraints.

\subsection{End-to-end Performance}

We conducted end-to-end training experiments on four models spanning different scales and architectures: LLaMA3.1-8B, LLaMA3.3-70B, Qwen3-30B-A3B, and Qwen3-235B-A22B, and measured training throughput in tokens per second. For \SysName, we enabled tensor parallelism (TP), FlashAttention kernels, and CUDA Graphs, with the tensor parallel size chosen according to the scale of the target model. The draft model was trained using ZeRO Stage 2 to achieve memory-efficient data-parallel execution. For the baseline, we evaluated both ZeRO Stage 2 and ZeRO Stage 3 configurations and report the best-performing setting.

Table~\ref{tab:e2e-perf} summarizes the results. \SysName consistently outperforms the baseline across all model scales, achieving a maximum speedup of 9.99×. The poor performance of the baseline can be attributed to two primary factors:
\begin{itemize}[leftmargin=*, itemsep=0pt, topsep=-7pt, parsep=0pt]
    \item Under ZeRO Stage 2, although gradients and optimizer states are sharded, the frozen target model parameters remain fully replicated on each device, leading to rapid scalability degradation as model size increases.
    \item ZeRO Stage 3 shards parameters, optimizer states, and gradients; however, frequent all-gather operations during target-model inference introduce substantial communication overhead, which severely limits throughput.
\end{itemize}

These results further highlight the effectiveness of target–draft decoupling. For large-scale models such as Qwen3-235B-A22B, ZeRO-style sharding leads to extremely low throughput due to communication-dominated execution. In contrast, \SysName consistently achieves strong performance by avoiding unnecessary synchronization and applying model-specific parallelization strategies. The results also underscore the importance of integrating with a mature inference engine like SGLang. As shown by the LLaMA-3.1-8B experiments, even when neither the baseline nor \SysName parallelizes the target model, \SysName still attains a 2.01× speedup, owing to the highly optimized prefill execution provided by SGLang.

\subsection{Impact of Target Model Backends}

In addition, we investigated the impact of different target model backends on training performance. In \SysName, we have supported three types of execution backends:

\begin{itemize}[leftmargin=*, itemsep=0pt, topsep=-7pt, parsep=0pt]
    \item \textbf{Hugging Face Backend}: Reuses model implementations from Hugging Face Transformers and relies on its internal \textit{tp\_plan} for tensor parallelism, when available.
    \item \textbf{SGLang Backend}: Reuses model implementations provided by SGLang and leverages its system-level optimizations, including chunked prefill~\cite{10.1145/3759441.3759444}, \texttt{torch.compile}, CUDA Graphs, and high-performance kernels~\cite{Ye2025FlashInferEA,10.5555/3737916.3740109}.
    \item \textbf{Custom Backend}: Includes models manually implemented by our team. This backend is particularly useful when a model is unavailable in Hugging Face Transformers or SGLang, or when the Hugging Face implementation lacks built-in parallelization support.
\end{itemize}

\begin{figure}[thb]
  \centering
  \includegraphics[width=0.9\linewidth]{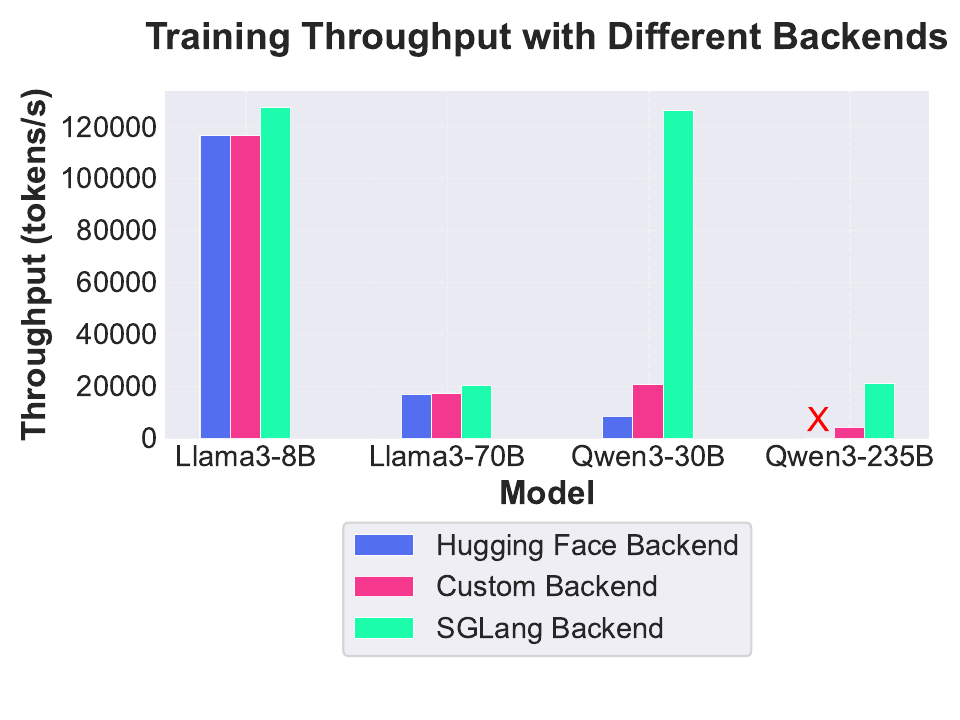}
  \caption{Training time with different execution backends}
  \label{fig:backend-perf}
\end{figure}

We conducted training experiments on the same set of models and tensor parallel configurations in Table~\ref{tab:e2e-perf}. As shown in Figure~\ref{fig:backend-perf}, SGLang significantly outperforms the other two execution backends, achieving speedups of up to 6.8×. These results highlight a key observation: optimizing the prefill stage is non-trivial, particularly for MoE models. For the Qwen3 experiments, both our custom backend and the Hugging Face backend exhibit substantially lower training throughput compared to SGLang. Notably, the Hugging Face implementation encounters runtime errors and fails to robustly support large-scale MoE models, further underscoring the importance of integrating with a mature, inference-optimized backend for scalable EAGLE3 training.

Another engineering advantage of integrating with SGLang is the clear separation of responsibilities. Model support and low-level inference optimizations can be delegated to the engine team, which typically adds support for newly released models promptly. This allows \SysName to focus on training-specific optimizations and system design, rather than duplicating model integration and maintenance efforts.

\subsection{Impact of Optimization Attention Kernel}

\begin{figure*}[thb]
  \centering
  \begin{subfigure}{0.45\textwidth}
    \centering
    \includegraphics[width=\linewidth]{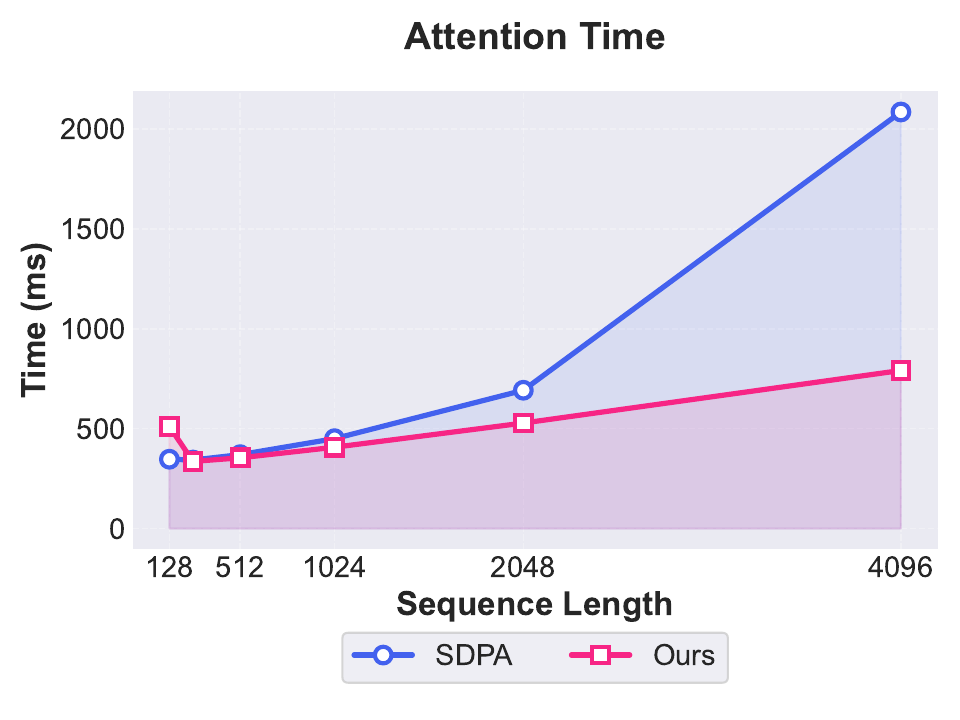}
        \caption{Kernel wall time for attention}
        \label{fig:attention-time}
  \end{subfigure}
  \hfill
  \begin{subfigure}{0.45\textwidth}
    \centering
    \includegraphics[width=\linewidth]{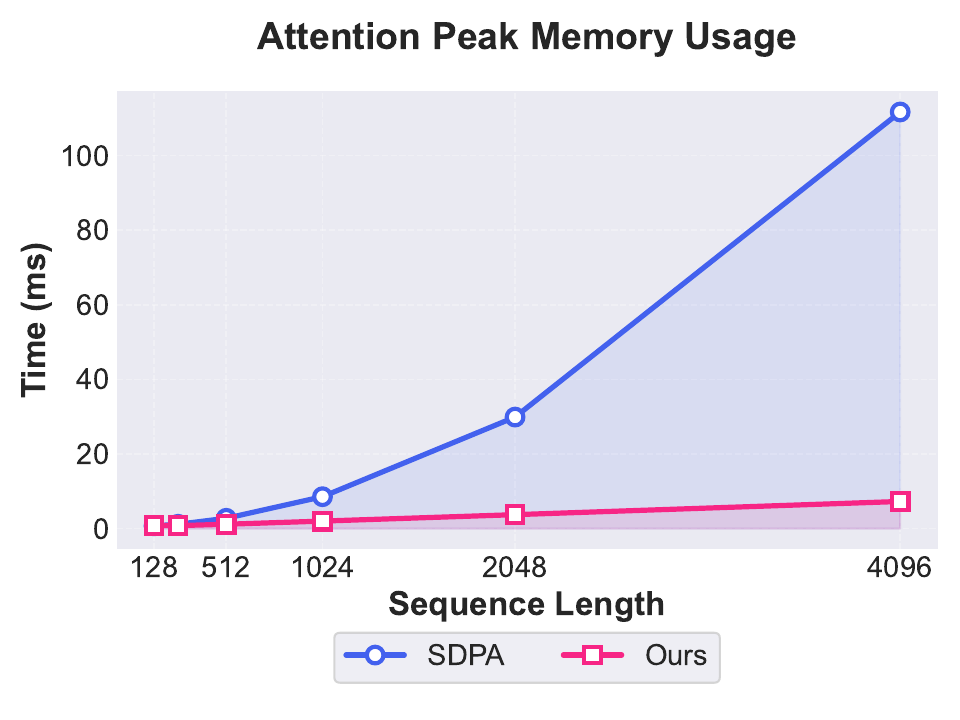}
    \caption{Peak memory consumption for attention}
    \label{fig:attention-mem}
  \end{subfigure}
  \caption{Comparison of execution time and memory usage between naive EAGLE3 attention and our optimized kernel.}
  \label{fig:attention-perf}
\end{figure*}

To evaluate the performance gains from our optimized attention kernel, we conducted micro-benchmarks comparing its execution time and peak memory usage against a native SDPA-based implementation. We set the TTT length to 7 and report measurements from the final TTT step. As shown in Figure~\ref{fig:attention-perf}, our optimized attention substantially reduces both wall-clock time and memory consumption. For a sequence length of 4096, it achieves reductions of 62.1\% in execution time and 93.5\% in peak memory usage on a single NVIDIA H200 GPU. Moreover, the performance gap widens as the sequence length increases, highlighting the effectiveness of our optimized kernel for training EAGLE3 under long-context settings.

\section{\BenchName}


As part of our open-source efforts, we trained the EAGLE3 draft models for a collection of mainstream open-source models including Llama, Qwen and Kimi. This collection is named \BenchName. 

We trained these on models on the Open-PerfectBlend dataset~\cite{xu2024perfectblendredefiningrlhf}, which consists of offers balanced 1.4M conversation in the chat, math, coding, instruction following domains. To achieve the best performance, we regenerated the assistant's responses in the dataset using the target model with temperature 0.8 and trained the model from scratch on the regenerated dataset. We trained the draft models for 2 epochs at learning rate 1e-4 with cosine annealing scheduler.

\begin{table*}[thb]
\centering
\small
\resizebox{0.8\linewidth}{!}{
\begin{tabular}{c|c|c|cc|cc|cc}
\toprule[1.5pt]
\multirow{2}*{\textbf{Target Model}} & \multirow{2}*{\textbf{Draft Model}} & \multirow{2}*{\textbf{\#GPUs}} & \multicolumn{2}{|c|}{\textbf{MTBench}} & \multicolumn{2}{|c|}{\textbf{GPQA}} & \multicolumn{2}{|c|}{\textbf{FinanceQA}} \\ \cline{4-9}
& & & \textbf{Throughput} & \textbf{Speedup} & \textbf{Throughput} & \textbf{Speedup} & \textbf{Throughput} & \textbf{Speedup} \\ \midrule[1pt]

\multirow{3}*{Llama-3.1-8B}     & -           & \multirow{3}*{1} & 190.0 & 1 & 190.5 & 1 & 185.7 & 1 \\
                                & Existing    &                  & \textbf{454.7} & \textbf{2.39} & 438.1 & 2.30 & 237.2 & 1.27 \\
                                & \BenchName        &                  & 450.0 & 2.37 & \textbf{514.2} & \textbf{2.70} & \textbf{258.6} & \textbf{1.39} \\ \hline
\multirow{3}*{Llama-3.3-70B}    & -           & \multirow{3}*{4} & 540.5 & 1 & 575.7 & 1 & 512.6 & 1 \\
                                & Existing    &                  & \textbf{1272.7} & \textbf{2.35} & 1049.0 & 1.82 & 981.7 & 1.92 \\
                                & \BenchName        &                  & 1253.0 & 2.31 & \textbf{1405.1} & \textbf{2.44} & \textbf{1022.7} & \textbf{2.00} \\
                                \hline
\multirow{3}*{Llama-4-Scout}    & -           & \multirow{3}*{8} & 502.1 & 1 & 541.0 & 1 & 288.9 & 1 \\
                                & Existing    &                  & 1253.0 & 2.50 & 1405.1 & 2.60 & 1022.7 & 3.54 \\
                                & \BenchName        &                  & \textbf{1312.4} & \textbf{2.61} & \textbf{1502.2} & \textbf{2.78} & \textbf{1189.6} & \textbf{4.12} \\
                                \hline
\multirow{2}*{Qwen-30B-A3B}     & -           & \multirow{3}*{4} & 1341.3 & 1 & 1410.4 & 1 & 1320.1 & 1 \\
                                & \BenchName        &                  & \textbf{2086.1} & \textbf{1.55} & \textbf{2341.3} & \textbf{1.66} & \textbf{1779.0} & \textbf{1.35} \\
                                \hline
\multirow{2}*{Qwen-235B-A22B}   & -           & \multirow{3}*{8} & 529.9 & 1 & 563.2 & 1 & 539.5 & 1 \\
                                & Existing    &                  & 642.7 & 1.21 & 716.7 & 1.27 & 689.4 & 1.28 \\
                                & \BenchName        &                  & \textbf{814.5} & \textbf{1.54} & \textbf{826.5} & \textbf{1.47} & \textbf{889.0} & \textbf{1.65} \\
                                \hline
\multirow{2}*{Ling-Flash-V2}    & -           & \multirow{2}*{8} & 728.5 & 1 & 794.1 & 1 & 747.7 & 1 \\
                                & \BenchName        &                  & \textbf{1022.6} & \textbf{1.40} & \textbf{1185.7} & \textbf{1.49} & \textbf{863.9} & \textbf{1.16} \\
                                \hline
\multirow{2}*{Kimi-K2}          & -           & \multirow{2}*{8} & 430.9 & 1 & 505.4 & 1 & 433.4 & 1 \\
                                & \BenchName        &                  & \textbf{533.8} & \textbf{1.24} & \textbf{811.4} & \textbf{1.61} & \textbf{660.0} & \textbf{1.52} \\
                                \bottomrule[1.5pt]
\end{tabular}}
\caption{Performance of various models on general benchmarks}
\label{tab:specbundle-general}
\end{table*}

\begin{table*}[t]
\centering
\resizebox{\linewidth}{!}{
\begin{tabular}{c|c|c|cc|cc|cc|cc}
\toprule[1.5pt]
\multirow{2}*{\textbf{Target Model}} & \multirow{2}*{\textbf{Draft Model}} & \multirow{2}*{\textbf{\#GPUs}} & \multicolumn{2}{|c|}{\textbf{LiveCodeBench}} & \multicolumn{2}{|c|}{\textbf{HumanEval}} & \multicolumn{2}{|c|}{\textbf{GSM8K}} & \multicolumn{2}{|c|}{\textbf{Math500}} \\ \cline{4-11}
& & & \textbf{Throughput} & \textbf{Speedup} & \textbf{Throughput} & \textbf{Speedup} & \textbf{Throughput} & \textbf{Speedup} & \textbf{Throughput} & \textbf{Speedup} \\ \midrule[1pt]

\multirow{3}*{Llama-3.1-8B}     & -           & \multirow{3}*{1} & 189.7 & 1 & 190.9 & 1 & 181.8 & 1 & 191.0 & 1 \\
                                & Existing    &                  & 398.4 & 2.10 & 480.3 & 2.52 & 228.6 & 1.26 & 422.4 & 2.21 \\
                                & \BenchName        &                  & \textbf{516.9} & \textbf{2.72} & \textbf{571.5} & \textbf{2.99} & \textbf{329.7} & \textbf{1.81} & \textbf{638.0} & \textbf{3.34} \\
                                \hline
\multirow{3}*{Llama-3.3-70B}    & -           & \multirow{3}*{4} & 560.9 & 1 & 561.0 & 1 & 453.2 & 1 & 567.4 & 1 \\
                                & Existing    &                  & 1303.4 & 2.32 & 1282.8 & 2.29 & 521.5 & 1.15 & 1122.2 & 1.98 \\
                                & \BenchName        &                  & \textbf{1459.4} & \textbf{2.60} & \textbf{1506.0} & \textbf{2.68} & \textbf{722.0} & \textbf{1.59} & \textbf{1524.9} & \textbf{2.69} \\
                                \hline
\multirow{3}*{Llama-4-Scout}    & -           & \multirow{3}*{8} & 484.3 & 1 & 631.9 & 1 & 455.9 & 1 & 561.8 & 1 \\
                                & Existing    &                  & 1601.3 & 3.31 & 1556.5 & 2.46 & 816.6 & 1.79 & 1479.0 & 2.63 \\
                                & \BenchName        &                  & \textbf{2170.2} & \textbf{4.48} & \textbf{1944.8} & \textbf{3.08} & \textbf{971.9} & \textbf{2.13} & \textbf{2110.3} & \textbf{3.76} \\
                                \hline
\multirow{2}*{Qwen-30B-A3B}     & -           & \multirow{3}*{4} & 1492.6 & 1 & 1366.6 & 1 & 1071.3 & 1 & 1469.0 & 1 \\
                                & \BenchName        &                  & \textbf{3413.0} & \textbf{2.29} & \textbf{3070.0} & \textbf{2.25} & \textbf{1499.6} & \textbf{1.40} & \textbf{3636.1} & \textbf{1.48} \\
                                \hline
\multirow{2}*{Qwen-235B-A22B}   & -           & \multirow{3}*{8} & 598.2  & 1 & 553.1 & 1 & 469.1 & 1 & 587.4 & 1 \\
                                & Existing    &                  & 803.8  & 1.34 & 889.9 & 1.61 & 697.0 & 1.49 & 821.8 & 1.39 \\
                                & \BenchName        &                  & \textbf{1155.7} & \textbf{1.93} & \textbf{1267.5} & \textbf{2.29} & \textbf{758.3} & \textbf{1.62} & \textbf{1399.2} & \textbf{2.38} \\
                                \hline
\multirow{2}*{Ling-Flash-V2}    & -           & \multirow{2}*{8} & 770.4  & 1 & 740.2 & 1 & 674.3 & 1 & 762.7 & 1 \\
                                & \BenchName        &                  & \textbf{1366.4} & \textbf{1.77} & \textbf{1359.0} & \textbf{1.83} & \textbf{1323.0} & \textbf{1.96} & \textbf{1685.6} & \textbf{2.21} \\
                                \hline
\multirow{2}*{Kimi-K2}          & -           & \multirow{2}*{8} & 500.1 & 1 & 466.1 & 1 & 337.9 & 1 & 492.1 & 1 \\
                                & \BenchName        &                  & \textbf{904.4} & \textbf{1.81} & \textbf{897.9} & \textbf{1.93} & \textbf{544.2} & \textbf{1.61} & \textbf{1022.7} & \textbf{2.08} \\
                                \bottomrule[1.5pt]
\end{tabular}}
\caption{Performance of various models on math and coding benchmarks}
\label{tab:specbundle-coding-math}
\end{table*}

\subsection{Evaluation Results}

We evaluated the results of \BenchName on a wide range of benchmark datasets:

\begin{enumerate}[leftmargin=*, itemsep=0pt, topsep=-7pt, parsep=0pt]
    \item Instruction-following: MTBench~\cite{Chen2025MTBenchAM}
    \item Math: Math500 and GSM8K~\cite{aime24}
    \item Coding: HumanEval~\cite{chen2021codex} and LCB~\cite{jain2024livecodebench}
    \item Other Subjects: GPQA~\cite{rein2024gpqa} and FinanceQA~\cite{mateega2025financeqabenchmarkevaluatingfinancial}
\end{enumerate}
We used SGLang as the inference engine to evaluate \BenchName models on the these benchmarks, with all experiments conducted on NVIDIA H200 GPUs. We compared our results against two baselines: (1) standard inference using a single target model and (2) speculative decoding with existing open-source draft models, where available. Several EAGLE3 draft checkpoints were provided by the authors of EAGLE3~\cite{li2025eagle} as well as the LMSYS team. Notably, the availability of speculative decoding draft models remains limited, as many target models do not yet have publicly accessible draft checkpoints. For all experiments, we fixed the number of concurrent requests to 8 for LLaMA-3.1-8B due to its smaller model size, and to 16 for larger models, applying tensor parallelism according to the target model scale. We evaluated multiple speculative decoding configurations, varying the number of speculative steps, top-
k, and the number of draft tokens, including (3, 1, 4), (5, 1, 6), (5, 3, 6), (7, 1, 8) and (7, 4, 8). We presented the highest throughput among all configurations. 

Table~\ref{tab:specbundle-general} shows the performance on the general benchmarks and Table~\ref{tab:specbundle-coding-math} shows the performs specifically on the coding and math benchmarks. It is evident that \BenchName significantly outperforms the baselines on all benchmarks and all dense and MoE models: the speedup can reach up to \textbf{4.48}×  compared to inference with no speculative decoding and \textbf{1.35}×  compared to inference with an existing draft model.

Particularly for the coding and mathematics benchmarks, \BenchName achieves speedups over baselines ranging from 1.61× to 4.48× (Table~\ref{tab:specbundle-coding-math}). This performance gap arises because existing checkpoints are primarily trained on the ShareGPT and UltraChat datasets, which contain limited coverage of math- and code-centric samples. These results underscore the critical role of data composition in training a well-balanced and high-performing draft model. However, this improvement is not uniform across all domains. A trade-off can be observed, as reflected in the slight decrease in MT-Bench performance for LLaMA-3 8B and 70B models.

\BenchName enriches the open-source ecosystem with a broader supply of draft models and delivers substantial performance improvements for production-grade inference. While the current release focuses on instruct models, we plan to extend support to reasoning models and vision–language models in future iterations.

\begin{figure*}[thb]
  \centering
  \begin{subfigure}{0.48\textwidth}
    \centering
    \includegraphics[width=\linewidth]{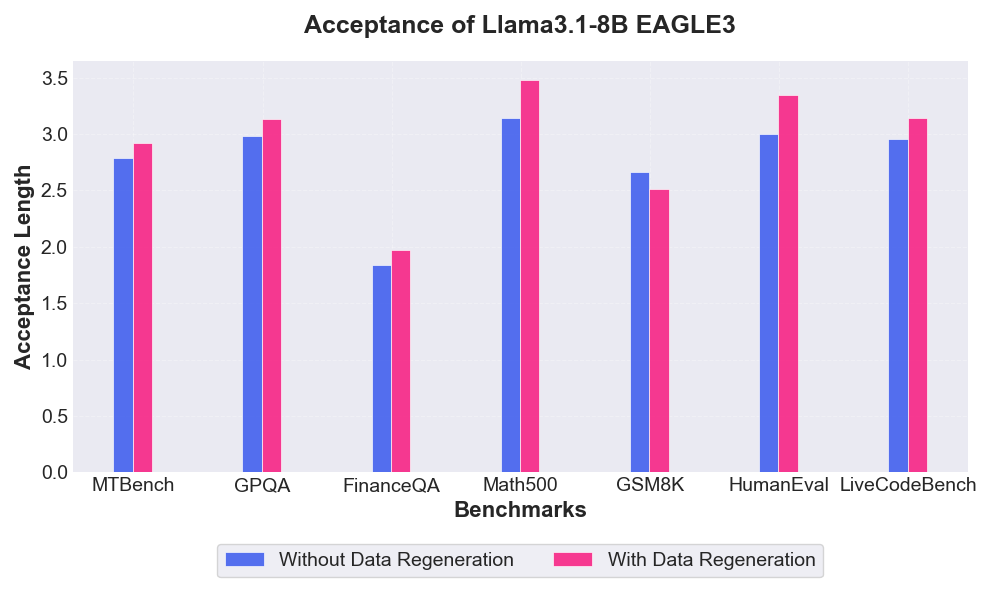}
    \caption{Acceptance Length}
    \label{fig:data-regen-acc}
  \end{subfigure}
  \hfill
  \begin{subfigure}{0.48\textwidth}
    \centering
    \includegraphics[width=\linewidth]{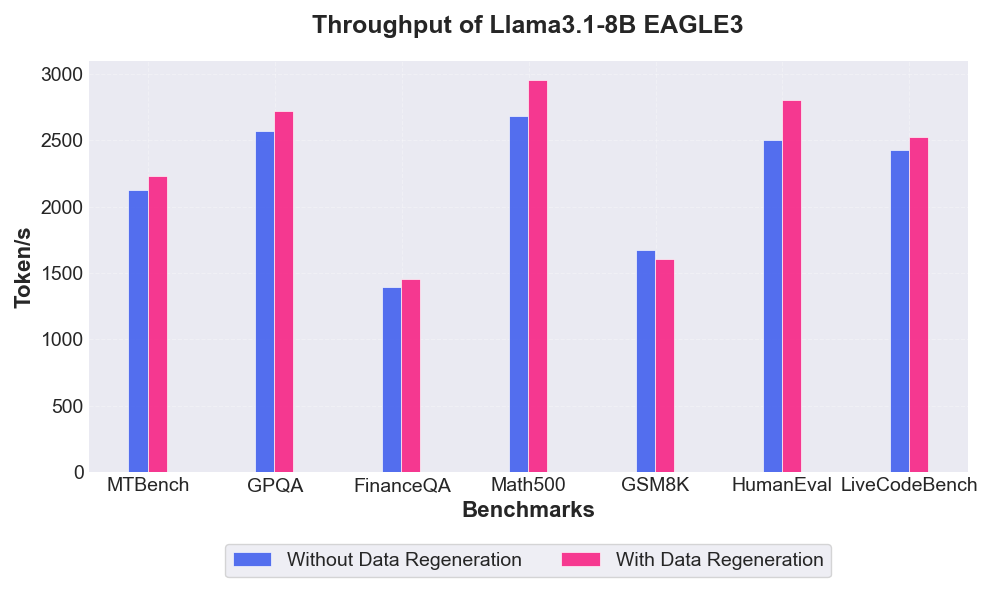}
    \caption{Output throughput}
    \label{fig:data-regen-throughput}
  \end{subfigure}
  \caption{Inference performance of Llama3.1-8B with EAGLE3 trained on datasets with and without regenerating the responses. The experiment was conducted on 1 H200 GPU with batch size 8.}
  \label{fig:inference-data-regen}
\end{figure*}

\section{Training Insights}

We draw some interesting insights for speculative decoding. 

\subsection{Impact of Data Regeneration}

Previous work claims that EAGLE methods exhibit low sensitivity to training data and therefore recommends training directly on the original dataset to reduce computational costs~\cite{Li2024EAGLESS}. However, our empirical results suggest that this assumption does not always hold. We trained an EAGLE3 draft model for LLaMA-3.1-8B using both the original PerfectBlend dataset and a regenerated version. As shown in Figure~\ref{fig:inference-data-regen}, data regeneration consistently increases the acceptance length across nearly all benchmarks, with FinanceQA being the only exception. Moreover, data regeneration yields an average throughput improvement of 5.3\% across all benchmarks.

Although the absolute throughput gain is moderate, data regeneration can have a substantial impact on long-term inference efficiency. Given that speculative decoding is widely deployed in online model serving systems such as the OpenAI API, even modest improvements can translate into significant reductions in inference cost at scale.

\begin{figure}[t]
  \centering
  \includegraphics[width=0.9\linewidth]{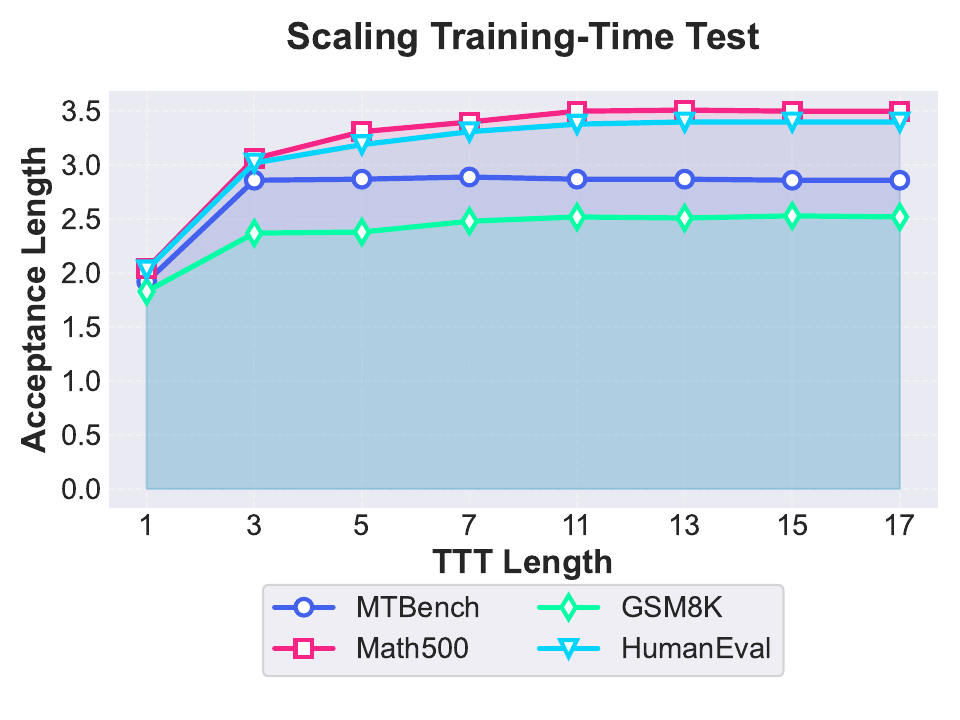}
  \caption{Scaling TTT for the Llama3.1-8B model on the perfect-blend dataset.}
  \label{fig:ttt-scaling}
\end{figure}

\begin{table*}[t]
\centering
\resizebox{\linewidth}{!}{
\begin{tabular}{c|c|c|c|c|c|c|c|c}
\toprule[1.5pt]
\textbf{Configuration}                                  & \textbf{\#Experts} & \textbf{MTBench} & \textbf{GPQA} & \textbf{FinanceQA} & \textbf{Math500} & \textbf{GSM8K} & \textbf{HumanEval} & \textbf{LiveCodeBench} \\ \midrule[1pt]
Same Params                                  & 2 & 1.16 & 1.05 & 1.07 & 1.17 & 1.05 & 1.2  & 1.14 \\ \hline
\multirow{3}*{Same FLOPS}                    & 2 & 1.61 & 1.56 & 1.34 & 1.67 & 1.59 & 1.64 & 1.46 \\
                                             & 3 & 1.62 & 1.69 & 1.35 & 1.77 & 1.7  & 1.67 & 1.57 \\
                                             & 4 & 1.63 & 1.7  & 1.36 & 1.79 & 1.72 & 1.69 & 1.59 \\ \hline
\multirow{3}*{With Shared Experts}& 2 & 1.52 & 1.69 & 1.5  & 1.75 & 1.61 & 1.53 & 1.52 \\
                                             & 3 & 1.52 & 1.7  & 1.46 & 1.75 & 1.61 & 1.53 & 1.52 \\
                                             & 4 & 1.53 & 1.69 & 1.47 & 1.81 & 1.67 & 1.64 & 1.54 \\ \hline
Dense Draft Model                            & - & 2.99 & 3.14 & 1.91 & 2.55 & 3.48 & 3.34 & 3.12 \\ \bottomrule[1.5pt]

\end{tabular}}
\caption{Acceptance rate of MoE models with different settings. The results are obtained with configurations MoE top-k = 1, EAGLE3 number of steps = 3, EAGLE3 top-k = 1 and EAGLE3 number of draft tokens = 4.}
\label{tab:moe-ablation}
\end{table*}

\subsection{Impact of Training-Time Test}

In the original implementation of EAGLE3~\cite{eagle-implementation, li2025eagle}, the TTT length is fixed at 7. We therefore conducted additional experiments to investigate the impact of TTT length on inference performance. Specifically, we varied the TTT length from 1 to 17. As shown in Figure~\ref{fig:ttt-scaling}, the results indicate that TTT is highly effective in improving the acceptance length, with a sharp gain observed as the TTT length increases from 1 to 3. Moreover, the optimal TTT length is task-dependent. For MT-Bench, a TTT length of 3 already achieves strong performance, whereas for more challenging and longer benchmarks, such as Math500, GSM8K, and HumanEval, a larger TTT length of approximately 13 yields the best results.

However, increasing the TTT length proportionally increases both training time and memory consumption for the draft model, introducing a clear trade-off between performance and efficiency. For domain-specific training, a practical strategy is to first conduct scaling experiments on a small subset of data to identify an appropriate TTT length before training on the full dataset.When training under limited resources, particularly memory constraints, it is advisable to reduce TTT to 3 or 5 to lower memory consumption. For cross-domain training, dynamically adjusting the TTT length based on the sample type could further reduce training cost, as not all samples require the same degree of training-time testing. We leave the design and evaluation of such dynamic TTT strategies to future work.

\subsection{Choice of Draft Models}

Recently released models such as LLaMA-4, DeepSeek-V3~\cite{DeepSeekAI2024DeepSeekV3TR}, and Kimi-K2~\cite{kimiteam2025kimik2openagentic} increasingly adopt the Mixture-of-Experts (MoE) architecture due to its superior performance and inference efficiency. However, existing EAGLE3 draft models remain dense. How to select the architecture of an appropriate draft model remains largely unexplored. 

Thus, we conducted experiments to investigate the suitability of MoE models as the draft model. We split the experiments into three categories:

\begin{itemize}[leftmargin=*, itemsep=0pt, topsep=-7pt, parsep=0pt]
    \item \textbf{Same Parameters}: We initialize two experts in the MoE layer of the draft model, with each expert using an intermediate dimension that is half that of the dense model. As a result, the combined parameter of the MoE layer matches that of the FFN layer in the dense draft model.
    \item \textbf{Same FLOPS}: We construct an MoE layer with two experts, where each expert has the same parameter count as the dense FFN layer. The number of experts selected per token is set to one, ensuring that the total number of floating-point operations remains unchanged.
    \item \textbf{MoE with shared experts}: On top of the "Same FLOPs" setting, we further introduce a shared expert. In this configuration, both the numbers of parameters and the total floating-point operations exceed those of the corresponding dense model.
\end{itemize}

The results are summarized in Table~\ref{tab:moe-ablation}. We observe that the dense draft model consistently outperforms all MoE variants across different settings, indicating that MoE draft models are inherently more difficult to train. Under the Same Params setting, the dense model’s FFN can be viewed as a degenerate two-expert MoE in which one expert has zero parameters. In contrast, the MoE model with the same total parameter budget performs poorly because each expert has fewer parameters than the dense FFN layer and therefore acts as a weaker learner.

Under the Same FLOPs setting, the MoE draft model performs noticeably better than in the Same Params case, as each expert has increased capacity and can learn more effectively. Nevertheless, its performance still lags behind that of the dense model. This gap arises because the routing top-k is set to 1, meaning that each expert is exposed to fewer tokens during training than the dense FFN, resulting in inferior generalization. Increasing the routing top-k could mitigate this issue, but also proportionally increase the per-token FLOPs, slowing down the drafting process. Consequently, despite their success as target models, MoE architectures are not well suited as draft models for speculative decoding.

\section{Conclusion}

In this paper, we presented \SysName, a highly efficient and scalable framework for training speculative decoding draft models, with first-class support for EAGLE3. We introduced target–draft decoupling and a set of optimized kernels that substantially reduce memory consumption and improve training throughput. Extensive experiments demonstrate that \SysName achieves up to 9.9× speedup over existing approaches. In addition, we released SpecBundle, a collection of production-grade, high-performance EAGLE3 draft models, and conducted systematic training analyses to distill practical insights that facilitate the real-world adoption of speculative decoding.

\nocite{langley00}

\clearpage
\bibliography{example_paper}
\bibliographystyle{icml2026}




\end{document}